\def\year{2021}\relax
\documentclass[letterpaper]{article} 
\usepackage{aaai21}  
\usepackage{times}  
\usepackage{helvet} 
\usepackage{courier}  
\usepackage[hyphens]{url}  
\usepackage{graphicx} 
\def\UrlFont{\rm}  
\usepackage{natbib}  
\usepackage{caption, subcaption} 
\usepackage{color}
\usepackage{amssymb}
\usepackage{amsbsy}
\usepackage{amsmath}
\usepackage{hyperref}
\usepackage{multirow}
\usepackage[ruled]{algorithm2e}
\usepackage{acronym}
\usepackage{xspace}
\usepackage{hyperref}
\usepackage{enumitem}
\usepackage{subcaption}
\usepackage[dvipsnames]{xcolor}
\usepackage{graphicx}
\usepackage{verbatim}
\usepackage[compact]{titlesec}
\usepackage{setspace}
\usepackage[]{algorithm2e}

\newcommand{\sign}{\text{sgn}}
\newcommand{\Hquad}{\hspace{0.3em}} 

\newtheorem{definition}{Definition}

\newcommand{\polygon}{\texttt{polygon}}
\newcommand{\edge}{\texttt{edge}}
\newcommand{\vertex}{\texttt{vertex}}
\acrodef{meip}[MEIP]{Maximum Entropy Inverse Planning}

\title{Generalized Inverse Planning: \\Learning Lifted non-Markovian Utility for Generalizable Task Representation}

\author{

    Sirui Xie\textsuperscript{*},
    Feng Gao\textsuperscript{*},
    Song-Chun Zhu
    \\
}
\affiliations{
    UCLA Center for Vision, Cognition, Learning and Autonomy\\
    srxie@ucla.edu, f.gao@ucla.edu, sczhu@stat.ucla.edu
}

\begin{document}

\maketitle

\begin{abstract}
 In searching for a generalizable representation of temporally extended tasks, we spot two necessary constituents: the utility needs to be \emph{non-Markovian} to transfer temporal relations invariant to a probability shift, the utility also needs to be \emph{lifted} to abstract out specific grounding objects. In this work, we study learning such utility from human demonstrations. While inverse reinforcement learning (IRL) has been accepted as a general framework of utility learning, its fundamental formulation is \emph{one} concrete Markov Decision Process. Thus the learned reward function does not specify the task independently of the environment. Going beyond that, we define a domain of generalization that spans \emph{a set of} planning problems following a schema. We hence propose a new quest, \emph{Generalized Inverse Planning}, for utility learning in this domain. We further outline a computational framework, \emph{Maximum Entropy Inverse Planning} (MEIP), that learns non-Markovian utility and associated concepts in a generative manner. The learned utility and concepts form a task representation that generalizes regardless of probability shift or structural change. Seeing that the proposed generalization problem has not been widely studied yet, we carefully define an evaluation protocol, with which we illustrate the effectiveness of MEIP on two proof-of-concept domains and one challenging task: \emph{learning to fold from demonstrations}.
\end{abstract}

\section{Introduction} 
Humans learn underlying utility by observing others' behaviors. It is widely accepted that we humans have a Theory of Mind (ToM), assume others as bounded rational agents, and inversely solve for their utility to understand their planned behaviors \cite{baker2009action}. The inferred utility is associated with some concepts, which together specify the task. Then in a similar context, this utility can generalize to incentivize us to act similarly. In this work, we study a formal definition of such generalization and a proper machinery to learn such utility.  

The utility we want to study is different from the reward function in classical reinforcement learning. In their seminal book, \citet{sutton1998reinforcement} distinguish planning from reinforcement learning as requiring some explicit \emph{deliberation} using a world model. It is this deliberation that the utility we discuss here expects to capture. 

\begin{figure}[h!]
    \centering
    \begin{subfigure}[b]{0.35\linewidth}
        \includegraphics[width=\linewidth]{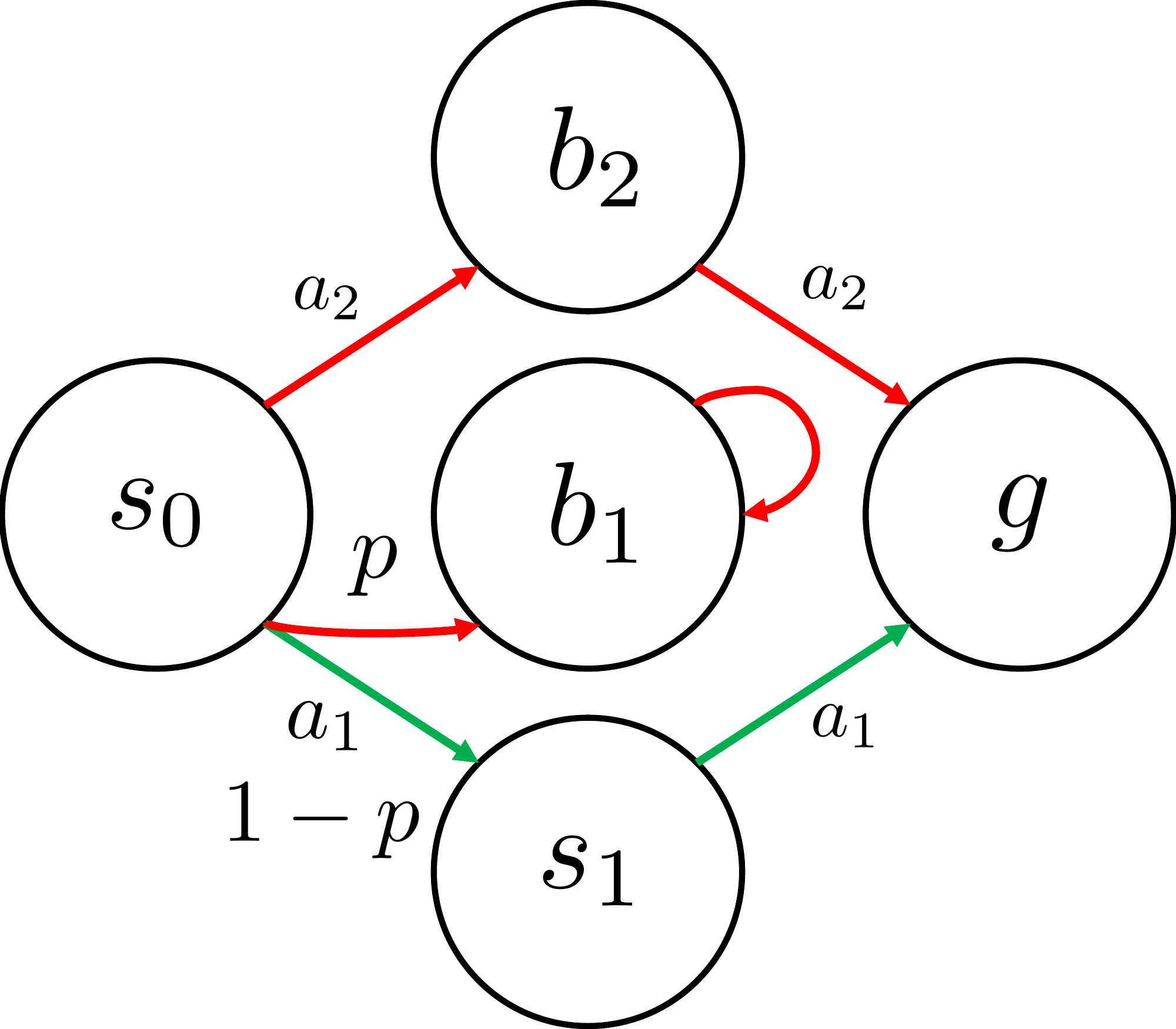}
        \caption{}
        \label{fig:shirt}
    \end{subfigure}
    \begin{subfigure}[b]{0.62\linewidth}
        \includegraphics[width=\linewidth]{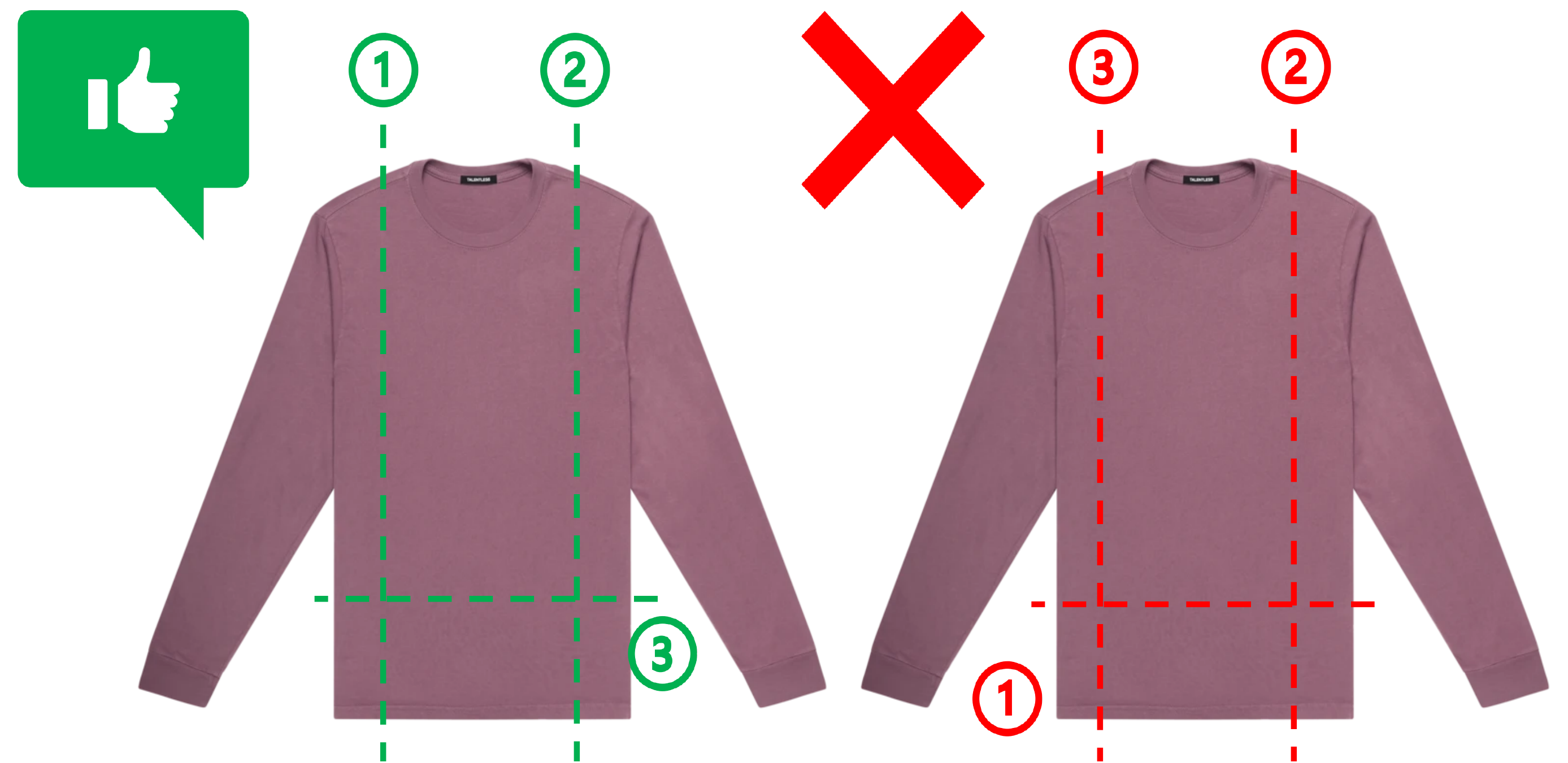}
        \caption{}
        \label{fig:fig1_mdp}
    \end{subfigure}
    \caption{(a) An MDP on which the agent needs to reach $g$ without hitting a $b$ (highlighted with the green arrow). Any Markovian rewards that make this expected behavior optimal are coupled with $p$. (b) The \emph{default} and \emph{weird} sequences for folding a cloth. The former should have higher utility. The underlying utility should also generalize robustly.}
    \label{fig:intro}
    \vspace{-1em}
\end{figure}

Apart from these philosophical concerns, there are also computational issues in the setup of classical reinforcement learning. One of our key insights is that tasks specified with Markovian reward functions do not generalize over environments. Consider a didactic example from \citet{littman2017environment}, see Fig.\ref{fig:intro}a. The desired behavior we want to specify is `maximizing the probability of reaching the \emph{goal} $g$ without hitting a \emph{bad state} $b$'. It is equivalent to a temporal description `\emph{do not} visit a $b$ \emph{until} you reach the $g$', which unfortunately cannot be represented with Markovian rewards independently from the environment. Concretely, let’s assume a discount of $\gamma = 0.8$ and a reward of +1 for reaching the goal. If $p = 0.1$, setting $r < -0.16$ to the bad state encourages the desired behavior. But if $p = 0.3$, this reward needs to be $r<-0.48$. Even though this example may seem contrived, it captures the essence of the limitation of Markovian rewards. There are more natural examples in our daily life. Imagine you are folding your clothes after laundry. Normally, \emph{not until} you fold the left and right sleeves \emph{do} you fold it in half from top to bottom, see Fig.\ref{fig:intro}b. Searching your memory, you will probably realize it is the \emph{default} order ever since you learned to fold clothes as a kid, no matter the one you fold is a T-shirt or a sweater, no matter it is your all-time favorite or a brand-new one. Remarkably, the utility you learned on this temporally extended task exhibits robust generalization. In other words, we have this cognitive capability to learn a task representation with a utility function that is independent of the environment. 

In fact, a slip in the \emph{transition probability} is only the mildest one in environmental shifts. In the example of cloth folding, \emph{hoodies} and \emph{T-shirts} have different \emph{structures} in the underlying probabilistic graphical models (PGM). Concretely, this is because they have different numbers of edges in their contours thus different numbers of nodes in their object-oriented probabilistic graphs. Despite of this difference, the utility learning mechanism ought to be tolerant of the heterogeneous nature of demonstrations. The learned utility should also help when folding a \emph{sweater} after seeing these demonstrations. From a classicist' perspective, this requirement goes beyond the formulation of RL, Markovian Decision Processes (MDP), where the language that specifies the structure of the MDP (which is essentially a PGM) is constrained to be \emph{propositional}. For readers who are not familiar with classical AI or computational linguistics, propositional logic can be understood as a language without object-orientation. Its object-oriented counterpart is \emph{first-order} or \emph{relational} logic. Intuitively, utility associated with a relation, \textit{i.e.,} ``symmetric'', can generalize better than than utility associated with a grounded description such as ``the left and right sleeves are symmetric''. In classical AI, this property is called \emph{lifted} in the sense of not being grounded with specific entities. 

How can machines learn utility that is both non-Markovian and lifted? The closest solution in the literature is Inverse Reinforcement Learning (IRL) \cite{abbeel2004apprenticeship}. However, IRL adopts the fundamental modeling of MDP. Given an MDP and a set of demonstrations from it, IRL learns a Markovian reward function by matching the \emph{mean statistics} of states or state-action pairs. This learned reward function can only encourage the expected behavior in the \emph{identical} MDP, because apparently it will fall into the trap of the didactic example above. Apart from that, utility learned with vanilla IRL also fails to generalize to a PGM with different structure. In this work, we propose a joint treatment for learning generalizable task representation from human demonstrations.

Our contributions are threefold: 
\begin{itemize}
    \item We characterize the domain of generalization in this utility learning problem by formally defining a \emph{schema} for the \emph{planning task} that represents a set of \emph{planning problems}. The target utility should be learned in a subset of this domain and successfully generalize to other problem instances. This is in stark contrast to the formulation of IRL for which only one planning problem is of interest. We hence term the problem \emph{generalized inverse planning}. 
    \item We propose an energy-based model (EBM) for generalized inverse planning. In Statistics, energy-based models are also dubbed \emph{descriptive models}. This is because the model is expected to match the minimal statistical description in data. It is this description that differentiates the proposed model, Maximum Entropy Inverse Planning (MEIP), from prior arts such as MaxEnt-IRL \cite{ziebart2008maximum}. Instead of matching the mean statistics in demonstrations under a Markovian assumption, MEIP matches the \emph{ordinal statistics}, a description that sufficiently captures the temporal relations in non-Markovian planning. This model can be learned with Maximum Likelihood, sampled with Monte Carlo Tree Search (MCTS). To combine MEIP with a first-order \emph{concept language}, we further introduce a boosting method to pursue abstract concepts associated with the utility. 
    \item Seeing that the generalization problem in this work was barely studied systematically, we carefully design an evaluation protocol. Under this protocol, we validate the generalizability of utility learned with MEIP in two proof-of-concept experiments for environmental change and a challenging task, \emph{learning to fold clothes}.
\end{itemize}

\section{Background}
\subsection{Inverse Reinforcement Learning}
Imitation learning, also known as learning from demonstrations, is a long-standing problem in the community of artificial intelligence. Earliest works most adopt the paradigm of Behavior Cloning (BC), directly supervise the policy at each step of provided sequences \cite{hayes1994robot}, \cite{amit2002learning}. \citet{atkeson1997robot} was the fist to consider the temporal drifting  in sequences. Nonetheless, BC is always believed to be less transferable without generatively modeling the decision making sequences. Alternatively, \citet{ng2000algorithms} proposed another paradigm, IRL, to inversely solve for the reward function in an MDP. The learned reward function is expected to help the agent learn a new policy, hopefully covers more states in the MDP. Together with some extensions such as \citet{abbeel2004apprenticeship} and \citet{ratliff2006maximum}, they set up the standard formulation of IRL. 

Consider a finite-horizon MDP, which is a tuple $\langle\mathcal{S}, \mathcal{A}, \mathcal{T}, R, h\rangle$, where $\mathcal{S}$ is the set of states, $\mathcal{A}$ is the set of actions, $\mathcal{T}: \mathcal{S} \times \mathcal{A} \times \mathcal{S} \rightarrow \mathbb{R}_{\geq 0}$ is the Markovian transition probability, $R: \mathcal{S} \times \mathcal{A} \rightarrow \mathbb{R}$ is the Markovian transition probability and $h$ is the horizon. \citet{abbeel2004apprenticeship} assume the Markovian reward to be linear to some predefined features $\phi(s)\in\mathbb{R}^k$ of states $s \in \mathcal{S}$ and derive the \emph{feature expectation} as $\mu\in\mathbb{R}^k$. Specifically, the underlying unknown reward is assumed to be $R^*(s)=\omega^*\cdot\phi(s)$ with unknown parameter $\omega^*\in\mathbb{R}^k$. Given a set of demonstration trajectories $\Psi^E=\{\zeta_1^E, \zeta_2^E, ..., \zeta_{m}^E\}$ from this MDP with $R^*(s)$, $\zeta\in{\mathcal{S}}^h$, the value with the estimated parameter $w$ is 
\begin{equation}
\label{eq:irl_value}
    \frac{1}{m}\sum_{i=1}^{m}\sum_{j=1}^{h} R(s_i^j) = 
    \frac{1}{m}\sum_{i=1}^{m}\omega\cdot\sum_{j=1}^{h}\phi(s_i^j) = \frac{1}{m}\sum_{i=1}^{m}\omega\cdot\mu(\zeta_i).
\end{equation}
The algorithm of IRL always runs with an off-the-shelf reinforcement learning method to generate trajectories $\Psi^G=\{\zeta_1, \zeta_2, ..., \zeta_{n}\}$ with the optimal value for the same MDP except that the reward $R(s)$ is the current estimation. We thus have two sets of trajectories. Assuming humans behave rationally when demonstrating, \citeauthor{abbeel2004apprenticeship} then propose a learning objective to maximize the margin between these two sets. When the model converges, the \emph{mean statistics} $\mu(\zeta)$ from $\Psi^E$ and $\Psi^G$ should be matched. 

To account for humans' bounded rationality and imperfect demonstration, \citet{ziebart2008maximum} introduced a probabilistic modeling for IRL. They started from a statistical mechanics perspective: the IRL model above does not address the ambiguity that many reward functions can lead to the same feature count. They propose MaxEnt-IRL to maximize the entropy of distribution of trajectories to break the tie while matching the mean statistics. The distribution of a trajectory $\zeta_i$ in a deterministic MDP is thus 
\begin{equation}
    P(\zeta_i | \omega)= \frac{1}{Z(\omega)} \exp(\omega\cdot\mu(\zeta_i))=\frac{1}{Z(\omega)} \exp(\omega\cdot\sum_{s_i^j\in\zeta_i}\phi(s_i^j)),
\label{eq:irl_dist}
\end{equation}
where $Z(\omega)$ is the partition function. For non-deterministic MDPs, there is an extra factor $P(\zeta_i)$ to account for the transition probability $\mathcal{T}$. 
In MaxEnt-IRL, the reward function is learned with Maximum Likelihood Estimate (MLE). Some variants such as \citet{boularias2011relative} and \citet{ortega2013thermodynamics} learn by minimizing relative entropy (KL divergence). 

We derive our model of Maximum Entropy Inverse Planning with the same principle. But in stark contrast to matching the mean statistics under the assumption of Markovian independence, MEIP matches the ordinal statistics \emph{Kendall rank correlation} to account for the non-Markovian temporal relations. To some degree, we share the same spirit with \citet{xu2009learning} and \citet{garrett2016learning}. But their models are discriminative, which are believed to be more data-hunger and less generalizable than \emph{analysis by synthesis} \cite{grenander1993general}.

\subsection{Structured Utility Function}

Prior attempts to make utility functions structured fall into two regimes: those that adopt First-order Logic (FOL) for lifting \cite{dvzeroski2001relational} \cite{kersting2008non} and those that adopt Linear Temporal Logic (LTL) for non-Markovian rewards \cite{li2017reinforcement},  \cite{littman2017environment}, \cite{icarte2018using}. Leveraging the expressive powers of language and axioms, these structured utility functions can be generalized over a \emph{task domain}. We draw inspiration from them. There are works that also inversely solve for structured utility function from demonstrations \cite{munzer2015inverse}, \cite{vazquez2018learning}, but they only focus on one regime. To the best of our knowledge, our work is the first to provide a unified probabilistic model for both regimes. Our framework adopts a relational language for abstract and composable concepts and match ordinal statistics with a descriptive model. The learned structured utility function is expected to represent a task in \emph{Generalized Planning}, to be elaborated below. 

\section{Generalized Inverse Planning}
\label{sec:prob_statement}
The planning we want to discuss in this work is generalized in the sense that its representation generalizes over a \emph{domain} with multiple \emph{planning problems} \cite{srivastava2011new}. It is thus fundamentally different from the formulation of reinforcement learning which only maximizes rewards on \emph{one} specific MDP. 
\begin{definition}[Planning Domain]
A planning domain $\mathcal{D}=\langle\mathcal{F}, \mathcal{A}\rangle$ consists of a set of fluents $\mathcal{F}$ which are real-valued functions and a set of actions $\mathcal{A}$ which are associated with some entities or parameters or both. 
\end{definition}
Obviously, fluents here are generalized predicates in classical symbolic planning, whose arity should be given in advance and value may vary with time. Actions may also be associated with preconditions and effects, specified by sets of fluent values. This definition makes the planning domain lifted to the first-order, abstracting out instances of objects, agents, and events, only describes their classes and relations. It ,therefore, generalizes to arbitrary numbers of instances. In the literature, STRIPS-style action language \cite{strips1971} provides such abstraction. 
\begin{definition}[Stochastic Planning Problem]
A stochastic planning problem $\Pi$ $= $ $\langle \mathcal{D}, \mathcal{E}, \mathcal{T}, R, \rho \rangle$ is given by a domain $\mathcal{D}$, a set of entities $\mathcal{E}$, a set of transition probabilities $\mathcal{T}$ for $\mathcal{A}$, a reward function $\mathcal{R}$ and a distribution $\rho$ over initial states. A state is specified by a set of fluent values. 
\end{definition}
Here we do not assume the transition probabilities $\mathcal{T}$ to be Markovian. It can thus be easily generalized to Partially Observable \cite{kaelbling1998planning} or semi-Markovian \cite{sutton1999between}. States are sets of fluent values, which may be grounded spatial relations, grounded temporal relations, or output of functions. The reward function $R$ is preferred over \texttt{Goal} because the extensionality of the latter one might be hard to specify, particularly for inversion from demonstrations. As introduced earlier, Markovian rewards may not generalize over a domain independently to the environment,  we assume $R$ to be non-Markovian for each planning problem. A \emph{plan} is a (sub)optimal trajectory of states for a planning problem. 

Given a set of plans from different planning problems in the same domain, a \emph{task} is their abstraction that generalizes these plans and captures their underlying optimality. It may abstract out numerical values in $R$. At the minimal level, it is the action or transition at each state that we care about. This principle was previously discussed in \citet{martin2004learning}, \citet{natarajan2011imitation}, and \citet{silver2020few}, which directly imitate the behavior with expressive policies whose hypothesis space is logically or programmatically constrained. But completely depending on the policy also constrained the capability of generalization. For example, the learned policy may not work or generalize well if the action set is (uncountably) infinite. To this end, an \emph{ordering} in the state space is still needed for a task to represent the (possibly bounded) rationality in demonstrated plans \cite{khardon1999learning}. 
\begin{definition}[Planning Task]
A planning task is a tuple $\langle\mathcal{D}, \prec\rangle$, where $\mathcal{D}$ is the domain and $\prec$ is a partial ordering relation over states. 
\end{definition}
This minimal algebraic description of a task for generalized planning. And \emph{Generalized Inverse Planning} is to inversely solve for a representation of the task that maintains this algebraic structure from demonstrations. 

Our proposal towards generalized inverse planning is a nested algorithm. In the inner loop, it learns a numerical representation of $\prec$ with maximum likelihood. In the outer loop it pursuits first-order concepts $\mathcal{F}$ with maximum a posteriori (MAP). We introduce from inside out in the next two sections and provide pseudo-code in Alg.\ref{alg:meip} and Alg.\ref{alg:pursuit}. 

\section{Maximum Entropy Inverse Planning}
\subsection{Descriptive Model and Maximum Entropy}
We adopt descriptive modeling for inverse planning. Different from a discriminative model, which models a conditional probability, \emph{i.e.} a classifier, a descriptive model specifies the probability distribution of the signal, based on an energy function defined on the signal through some descriptive feature statistics extracted from the signal \cite{wu2019tale}. In the literature of modern AI, it is also known as energy-based model \cite{lecun2006tutorial}. From the discussion above, it comes clear to us that given a set of plans from different problems in the same domain, the minimum statistical description to match should be ordinal to account for temporal relations. Therefore, different from the feature expectation in IRL, we match the Kendall ranking statistics for inverse planning given $n$ plans from the same domain:
\begin{equation}
    \tau = \frac{1}{n}\sum_{i=1}^{n}\frac{2}{(m_i)(m_i-1)}\sum_{j=1}^{m_i}\sum_{k=j+1}^{m_i}d(s^{j}_{i}, s^{k}_{i}),
\end{equation}
where $d(s^{j}_{i}, s^{k}_{i})$ scores the concordance or discordance of a ranking function $g$ for temporally indexed states $s_{j}^{i}$ and $s_{k}^{i}$ where $j<k$ from a plan $\zeta_i$ for a problem $\Pi_i$:
\begin{equation}
    d(s^{j}_{i}, s^{k}_{i})=\begin{cases}
    +1 & g(s^{k}_{i})-g(s^{j}_{i})>0\\
    -1 & g(s^{k}_{i})-g(s^{j}_{i})<0\\
    0 & g(s^{k}_{i})-g(s^{j}_{i})=0
    \end{cases}.
    \label{eq:kendall_tau}
\end{equation}

The range of Kendall $\tau$ is $[-1,1]$, where $\tau=1$ indicates a perfect match. However, normally there can be multiple distributions of plans and thus multiple ranking functions $g$ that match this ordinal statistics. Similar to MaxEnt-IRL, we employ the principle of maximum entropy \cite{jaynes1957information} to resolve ambiguities in choosing distributions. Concretely, we maximize the entropy of the distribution over plans under the constraint that the Kendall $\tau$ can be matched between the demonstrations and generated plans: 
\begin{equation}
\begin{split}
    \max \sum\nolimits_{\zeta} - P(\zeta | g) \log P(\zeta | g)\\
    s.t. |P(\zeta|g) \tau(\zeta) - 1|<\epsilon, \\
    \sum\nolimits_{\zeta} P(\zeta|g)=1, \quad P(\zeta|g) \geq0.
\label{eq:constrained}
\end{split}
\end{equation}
Under the KKT condition, we can derive the Boltzmann form of this distribution from Eq.\ref{eq:constrained}'s Langrangian:
\begin{equation}
    P(\zeta | g, \lambda)= \frac{1}{Z(g, \lambda)} \exp(\lambda\cdot\tau(\zeta))P(\zeta),
\label{eq:meip_dist}
\end{equation}
where $Z(g, \lambda)=\sum_{\zeta} \exp(\lambda\cdot\tau(\zeta))P(\zeta)$ is the partition function, $\lambda$ is the Langrangian multiplier and $P(\zeta)$ accounts for the the transition probability $\mathcal{T}$.   
\subsection{Utility Learning for Order Preserving}
Comparing Eq.\ref{eq:meip_dist} with Eq.\ref{eq:irl_dist}, it is easy to see the correspondence between $\lambda\cdot\tau(\zeta)$ and $\sum_{s^j\in\zeta}R(s^j)$. To further derive the utility function of plans from Eq.\ref{eq:meip_dist}, we first specify the concrete form of $g$. We assume for each state $s$, which is a set of grounded fluent values, there is a vector of fist-order concepts $f(s)$ that generalizes over the domain $\mathcal{D}$. We will introduce how this vector $f(s)$ can be learned in next section. Here we can simply assume it is given. And we can further assume that the ranking function $g(s)$ is piece-wise linear with respect to this concept vector $f(s)$. Being piece-wise linear is a general assumption because functions with this characteristics are in theory as expressive as artificial neural networks. Specifically, we discretize each entry $f_i(s)$ into $M$ bins and attach a vector $\omega_i$ as a functional:
\begin{equation}
\label{eq:gf}
    g_i(s) = \omega_i \cdot f_i(s) = \sum\nolimits_{j=1}^M \omega_i^j f_i^j(s).
\end{equation}
Obviously, there is no need to separate $\lambda$ and $\omega$ anymore. So we drop $\lambda$ for the derivation below. 

To further illustrate the utility function in MEIP, we can rewrite $\tau(\zeta)$ as:
\begin{equation}
    \tau(\zeta) = \sum_{i=2}^{h}\frac{2}{(h)(h-1)}\sum_{j<i}d(s^{j}, s^{i})=\sum_{i=2}^{h}R(s^i),
\label{eq:meip_tau}
\end{equation}
Therefore we have
\begin{equation}
    R(s^i) = \frac{2}{(h)(h-1)}\sum_{j<i} d(s^j, s^i) \propto \sum_{j<i} \sign(g(s^j)-g(s^i)).
\label{eq:meip_rew}
\end{equation}
It is easy to see that this reward function is non-Markovian, in stark contrast to $R(s)=\omega\cdot\phi(s)$ in MaxEnt-IRL. 

We can solve for the $\omega$ by maximum likelihood (MLE) over given demonstrations, which according to \citet{jaynes1957information} implies maximum entropy:
\begin{equation}
    \mathcal{L}_{\omega} = \frac{1}{n}\sum\nolimits_i\log P(\zeta_i | \omega) =\frac{1}{n}\sum\nolimits_i \tau(\zeta_i) -\log Z(\omega).
\label{eq:meip_loss}
\end{equation}
Notice that $\log P(\zeta_i)$ is dropped as a constant. If $\tau$ is differentiable \emph{w.r.t.} $\omega$, consider Eq.\ref{eq:meip_loss}'s gradient
\begin{equation}
\begin{split}
    \nabla_\omega \mathcal{L}_{\omega}  &=\frac{1}{n}\sum\nolimits_i \nabla_\omega\tau(\zeta_i) -\nabla_\omega\log Z(\omega)\\
    &= \frac{1}{n}\sum\nolimits_i \nabla_\omega\tau(\zeta_i) - \sum\nolimits_j P(\zeta_j | \omega) \nabla_\omega\tau(\zeta_j)\\
    &= \frac{1}{n}\sum\nolimits_i \nabla_\omega\tau(\zeta_i) - \mathbb{E}_{P(\zeta_j | \omega)}[\nabla_\omega\tau(\zeta_j)], 
\label{eq:meip_grad}
\end{split}
\end{equation}
the second term $\mathbb{E}_{P(\zeta_j | \omega)}[\nabla_\omega\tau(\zeta_j)]$ can be approximated by sampling. Apparently, there is an contrastive view here: when maximizing the likelihood of demonstrations, we maximize their Kendall $\tau$ and minimize the Kendall $\tau$ of generated plans. The underlying intuition is that at convergence, for pairs of states $\langle s_i^m, s_i^n \rangle$ in the demonstrations, $g(s_i^m)<g(s_i^n)$ if $m<n$; for pairs $\langle s_j^m, s_j^n \rangle$ in generated plans, $g(s_j^m)\geq g(s_j^n)$ if $m<n$; for pairs with one state $s_i^m$ in demonstrations and the other one $s_j^m$ from generated plans, $g(s_i^m)\geq g(s_j^m)$. We would refer to ordinal relations listed here as $\hat{d_k}$, resembling groundtruth labels. 

However, $\tau$ is not differentiable. To this end, we need to learn a classifier with $\langle g(s^m), g(s^n)\rangle$ to match the order described above, mimicking the optimization in Eq.\ref{eq:meip_grad}. One way is to directly relax $d(s^{m}, s^{n})$ with a discriminator $D(s^{m}, s^{n})=\tanh(g(s^n)- g(s^m))$. To some extent, this approximation shares similar spirits with Generative Adversarial Imitation Learning (GAIL) \cite{ho2016generative} and some variants \cite{finn2016connection}. But different from them, we explicitly consider the temporal relation in the non-Markovian utility. 

We can also learn this classifier with max-margin methods. Essentially, it is a Ranking SVM model \cite{liu2011learning} taking both the demonstrations $\Psi^E$ and the sampled plans $\Psi$ as supervision. Consider there are $K$ pairs of states $\langle s_k^m, s_k^n\rangle$ in total from $\Psi^E\lor\Psi$ expected to fulfill the ordinal relation $\hat{d_k}$ above, we have this Quadratic Optimization problem:
\begin{equation}
\label{eq:meip_svm}
\begin{split}
    &\min_{\omega} \mathcal{L}_\omega^{SVM}=\frac{1}{2}\left\lVert w\right\rVert^2+ C \sum_k^K\xi_k, \quad s.t.\\
    &\forall k: \xi_k \geq 0, \hat{d_k} \cdot(g(s_k^n)-g(s_k^m)) \geq 1-\xi_k.
\end{split}
\end{equation}
It can be solved by off-the-shelf Gradient Descent (GD) or Quadratic Programming (QP). Notice that the number of constraints and slack variables grows quadratically in the size of each plan, we thus only consider the ordinal relations in each planning problem $\Pi$. But the parameters $\omega$ are shared for all problems in the domain $\mathcal{D}$. If we only consider the linear, primal form of the problem, there are also efficient methods for training \cite{joachims2006training}.
\subsection{Sampling Method}
We need to sample from the EBM to calculate parts in Eq.\ref{eq:meip_svm} that correspond to the second term in Eq.\ref{eq:meip_grad}. Thus we need to have the distribution of trajectories from the EBM. In this work, we only consider planning problems where states are discrete. We leave the continuous state space as our future work. We can factorize the probability of a plan $\zeta_i=(s_i^0, s_i^1, ..., s_i^h)$ by conditioning:
\begin{equation}
    P(\zeta_i | \omega) = P(s_i^0)P(s_i^1|s_i^0,\omega)P(s_i^2|s_i^1, s_i^0,\omega)...
\end{equation}
We take its logarithmic form to connect with Eq.\ref{eq:meip_tau} and Eq.\ref{eq:meip_rew}:
\begin{equation}
\begin{split}
    \log P(\zeta_i | \omega) &= \log \rho(s_i^0)+\sum\nolimits_{j=1}^{h}\log P(s_i^j|s_i^0...s_i^{j-1},\omega),   
\end{split}
\end{equation}
where $\log P(s_i^j|s_i^0...s_i^{j-1},\omega)$ takes both the action and transition probability into account. Since here we do not explicitly specify the action in $P(\zeta_i | \omega)$\footnote{The main concern here is that grounded actions in plans from different $\Pi$ may not generalize over $\mathcal{D}$. \citet{bonet2018features} discuss learning abstract actions for generalized planning. Here we do not enforce a relational form in the action space $\mathcal{A}$. We assume the equivalence between actions causing same transitions. We further assume there is an absorbing state \emph{failure} for each state transition thus $\mathcal{T}$ is stochastic with $P(s_i^j|a_i^j,s_i^0...s_i^{j-1})\neq1$.}, we decompose it to be
\begin{equation}
    \log P(s_i^j|s_i^0...s_i^{j-1},\omega) = R_\omega(s_i^j) + \log P(s_i^j|s_i^0...s_i^{j-1}).
\end{equation}
It then becomes clear that to sample from the descriptive model, we just sample proportionally to $\exp(\sum_j R_\omega(s_i^j))$:
 \begin{equation}
    P(\zeta_i|\omega)=P(s_i^0...s_i^j...s_j^h|\omega) \propto \exp(\sum\nolimits_j R_\omega(s_i^j)).
\end{equation}
$\sum\nolimits_j R_\omega(s_i^j)$ can be acquired from Monte Carlo Tree Search, for which the reward at each node is initialized with $R_\omega(s_i^j)$. After its convergence, we can sample trajectories according to the value of each branch. 

It is worth noticing that the same sampling method can be used to plan for optimal utility when transferred to other problems $\Pi_{new}$ in the same domain $\mathcal{D}$. So generally speaking, for sampled plans, there is a min-max view:
\begin{equation}
    \min\nolimits_\omega \max\nolimits_\zeta \mathbb{E}_{P(\zeta | \omega)}[\tau(\zeta)]. 
\end{equation}

\begin{algorithm}[t]
\textbf{Input:} A Set of Concepts $\Delta$ from the Concept Language; MCTS Convergence Conditions; Hyperparams.\\
\KwData{Human Demonstrations $\Psi^E=\{\zeta_i\}$}
\KwResult{Learned Parameters of Utility $\omega$}
\textbf{Init:} $\omega$, value function $V$ in MCTS\\
\While {utility function not converged} {
    \While{MCTS not stopped} {
            MCTS rollout, generate $\zeta$\\
            Compute Kendall $\tau_\omega(\zeta)$ and $R(s)$ with Eq. 8\\
            Value iteration via MCTS with Kendall $\tau_\omega$
    }
    sample trajectories $\Psi=\{\zeta_j\}$ with MCTS according to converged values of branches\\
    update $\omega$ with $\Psi^E$ and $\Psi$ (See Eq. 11)
}
\caption{Maximum Entropy Inverse Planning}
\label{alg:meip}
\end{algorithm}

\section{Learning First-Order Concepts}
In the previous section, we introduced a descriptive model for inverse planning, given some concepts that generalize over the planning domain. In this section, we introduce a formalism for concepts with this characteristics and how we learn them as $\mathcal{F}$ in Generalized Inverse Planning. 

\subsection{Concept Language}
As introduced in Sec.3\ref{sec:prob_statement}, a planning domain $\mathcal{D}$ is defined in a lifted manner, abstract out instances of entities in specific problems. Therefore the computational form of the utility learned from Generalized Inverse Planning should also be lifted and be invariant to the variation of numbers of instances. 
In AI, the first-order logic with quantifiers and aggregators is a formalism to express this invariance. To this end, we employ a modified concept language \cite{donini1997complexity} as a grammar for elements in $\mathcal{F}$ such that we can learn these first-order concepts in a top-down manner. Concept languages have the expressive power of subsets of standard first-order logic yet with a syntax that is suited for representing \emph{classes} of entities. Adopting the terminology of FOL, each concept is represented by a first-order \emph{formula}. However, in the original concept languages, concepts are assumed independent, which might not be the case for our utility function\footnote{Recall that we assume the ranking function $g(s)$ to be (piece-wise) linear in Eq.\ref{eq:gf}, which requires concepts $f_i(s)$ to be independent from each other. Utility functions whose concepts are not independent will never be expressed without this modification.}. So the primary modification we introduce is to complete it as in FOL, explicitly accounting for formulas which take other formulas as \emph{terms} 
with a syntax:
\begin{align*}
    C &\rightarrow C_a \mid P( C',C'',...) \mid F( C',C'',...),
\end{align*}
where the $'$ notation highlights different concepts. $P$ denotes predicates with binary value domains i.e. relations, $F$ denotes functions with either real or discrete value domains.\footnote{They may have neural network equivalents structuralized as Multi-Layer Perceptrons (MLP) or Graph Neural Nets (GNN).} They all have their own arities. $C_{a}\in\mathcal{C}_{a}$ are concepts represented by \emph{atomic formulas} with a syntax: 
\begin{align*}
    &C_a \rightarrow QV \mid AV, \Hquad V \rightarrow FD \mid PD, \Hquad D\rightarrow \text{ext}(P)\mid U,\\
    &Q \rightarrow \forall \mid \exists \mid \#, \Hquad A \rightarrow \max \mid \min \mid \text{avg},\\
    &P \rightarrow P_p \mid \neg P' \mid P' \land P'' \mid P^p \mid P^*.
\end{align*}
$A$ is the set of aggregators and $Q$ is the set of quantifiers. $V$ is a set that is either the value domain of a fluent $F$ or the truth domain of a predicate $P$. $D$ is the domain of interest that can either be the extension of a certain predicate, which is denoted as $\text{ext}(P)$, or the universe ($U$) of entities. The dimension of the domain $D$ should match the arity of $F$ and $P$ when placed together in $FD$ and $PD$. Among predicates $P$, $P_p$ are \emph{primitives}. Other predicates can be their constituents' negation ($\neg P'$) or conjunction ($P' \land P''$). They can also be a result of permuting another predicate's arguments ($P^p$), if the arity is larger than 1 and arguments are from the same class. They can even be defined transitively ($P^*$), as in original concept languages. Apparently, $\mathcal{C}_a\subset\mathcal{C}\subset \mathbb{R}$. 

\subsection{Concept Pursuit}
To consider the combination over the full bank of concepts $\mathcal{C}$ would be computationally intractable. A Bayesian treatment of concept induction can provide a principled way to incorporate the prior of Occam’s razor to probabilistic grammars. Let us denote the selected $K$ concepts with $\Delta_k=\{f_0, f_1,...,f_{k-1}\}$, where $f_i$ coincides Eq.\ref{eq:gf}, the posterior of the utility function $\langle \Delta_k, \Omega_k \rangle$ given demonstrated plans $\Psi=\{\zeta_0, \zeta_1,... \zeta_{n-1}\}$ would be
\begin{equation}
    P(\Delta_k, \Omega_k | \Psi) = \frac{P(\Psi|\Delta_k, \Omega_k)P(\Delta_k)P(\Omega_k)}{P_{k}(\Psi)}.
\end{equation}
To obtain the MAP estimate efficiently, we adopt stepwise greedy search over $\mathcal{C}$:
\begin{equation}
\label{eq:pursuit}
\begin{split}
    &C_{+}, \omega_{+} =\arg\max \log P(\Delta_{+}, \Omega_{+} | \Psi)-\log P(\Delta_{k}, \Omega_{k} | \Psi) \\
   &\approx \arg\max KL(P_{+}(\zeta)|P_{k}(\zeta)) + \log P(C)-\log P_{+}(\Psi)/P_{k}(\Psi)\\
   &\approx \arg\max |\mathbb{E}_\Psi [\tau_{C,\omega}(\zeta_i)]-\mathbb{E}_{P_{+}} [\tau_{C,\omega}(\zeta_j)]| + \log P(C).
\end{split}
\end{equation}
Specifically, concepts in $\mathcal{C}$ are first sorted by their complexity, reflecting the prior $P(C)$. Note that concepts are mutually exclusive if they share the same $V$ and only differentiate at $A$ or $Q$, so they are stored in the same slot in the sorted list. The levels of complexity are naturally discretized by the number of \emph{primitive} fluents or predicates involved. We start from the simplest level. At each level, the concept that brings the largest margin is added to $\Delta$ in a step-wise greedy manner. The selection terminates at the current level when the marginal benefit in the posterior is below a threshold. Then we move on to the next level. Since some complex concepts may have information overlap with simpler ones \textit{e.g.} $\forall (P_1\land P_2) D$ vs $\forall P_1 D \lor \forall P_2 D$, the simpler ones are replaced from $\Delta$ when the complex are added. The first term in Eq.\ref{eq:pursuit} is equivalent to the MLE in Eq.\ref{eq:meip_grad} and can be solved with MEIP. Therefore the complete algorithm is nested.

This derivation leads us to a boosting method \cite{friedman2001greedy}. A similar greedy search strategy was proposed for feature selection in IRL by \citet{bagnell2007boosting}. But different from them, we further assume the increase from $\log P(C)$ is always more significant than the former term, such that we only need to consider a subset of $\mathcal{C}$ with the lowest complexity at a time. 

\section{Experiments}
\subsection{Evaluation Protocol}
To help readers better understand the generalization problem studied in this work, we provide a systematic introduction of our evaluation protocol. Note that even though it is the \emph{learned utility} that is to be evaluated, it cannot be evaluated without planned behaviors. This because the optimal utility can not be trivially defined for most tasks where Generalized Inverse Planning is meaningful, a issue that originally motivated the proposal of IRL \citep{abbeel2004apprenticeship} 

\begin{enumerate}[noitemsep, nolistsep]
    \item The agent first learns the utility in the environment where the demos come from;
    \item We then transfer this agent to another environment, which is built under the same \emph{schema} thus in the same \emph{planning task}. This new environment is a result of probability shift or structural change or both;
    \item Given the symbolic world model of the new environment, the agent optimizes for a policy that maximizes the reward function with model-based methods such as MCTS;
    \item Test if the behavior of the agent in this new environment is consistent with the demo in terms of temporal relations.
\end{enumerate}
We adopt the following two sets of evaluation metrics, depending on the diversity of the desired behavior:
\begin{enumerate}[noitemsep, nolistsep]
    \item For simple tasks the ground-truth behavior is only one single sequence, such as in the didactic example $S_0\rightarrow S_1\rightarrow g$ is the only desired sequence, we evaluate the learned utility by a Monte Carlo estimate of the probability of the desired sequence executed by an MCTS agent's planned behavior in the new environment. 
    \item Most of the time, the ground-truth optimal utility is not clear to us, especially its numerical value. But we can extract the ground-truth concepts and their ordering from the demos. We evaluate the learned utility with the objective of IRL or inverse planning: the matching in the statistics. Specifically, given the planned behaviors and the demonstrated ones, we measure their \emph{mean matching} to check if the learned utility can attain a \emph{similar} behavior from the perspective of IRL; We measure their \emph{Kendall $\tau$} (ordinal matching) to check if the learned utility can attain a \emph{similar} behavior from the perspective of inverse planning. 
\end{enumerate}
\subsection{Experiment 1: Probability Shift}

\begin{figure}[t!]
    \centering
    \includegraphics[width=0.80\linewidth]{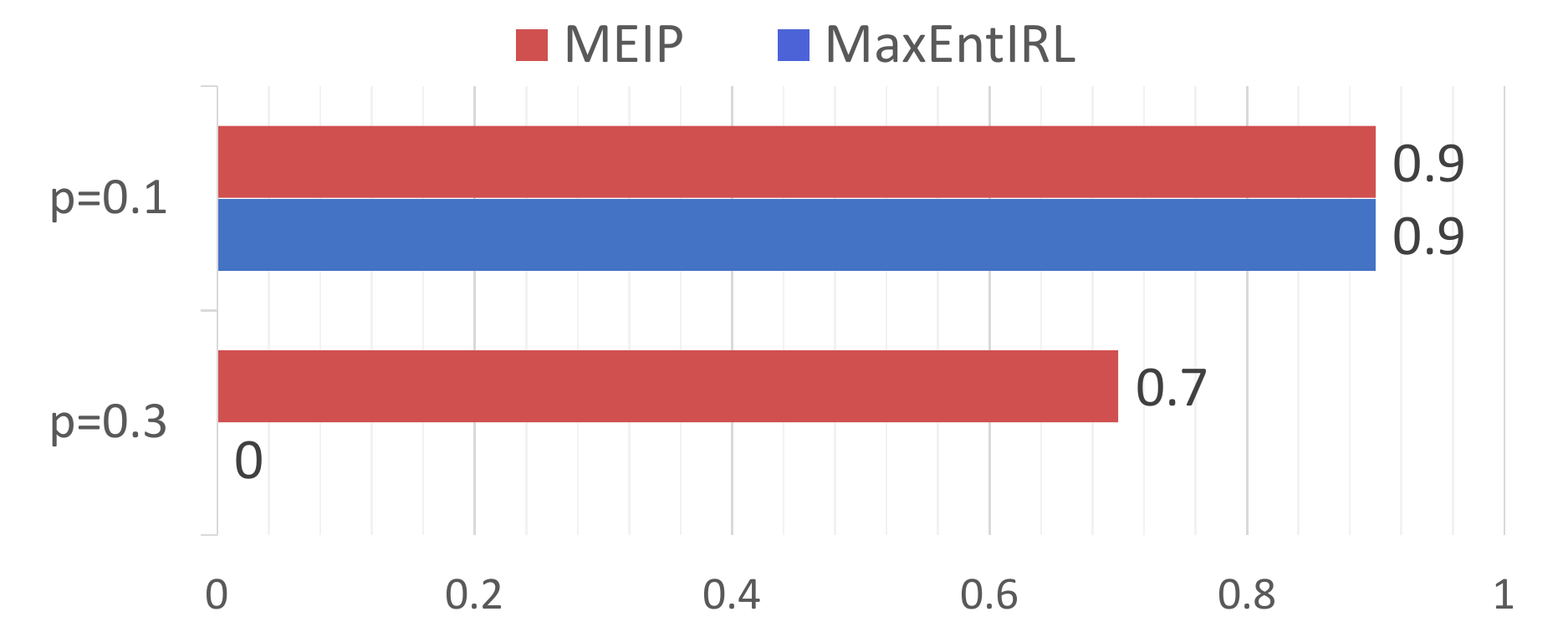}
    \caption{The probability of `reaching the goal state without hitting a bad state' in the training environment ($p=0.1$) and the testing ($p=0.3$) for agents whose utility is learned with MEIP and MaxEnt-IRL respectively. $1-p$ is the optimal.}
    \label{fig:prob_shift}
\end{figure}

We first conduct an experiment with the didactic example from \cite{littman2017environment} introduced above to illustrate utility learned with MEIP can generalize regardless of probability shift. Here probability shift refers to a change in a distribution in $\mathcal{T}$. The desired behavior is `\emph{do not} visit a bad state \emph{until} you reach the goal'. We recommend readers to review Fig.\ref{fig:intro}a for the problem setup. When learning the utility, the agent is provided with demonstrations collected from the MDP with $p=0.1$. And it is also provided a black-box model to simulate this MDP during MCTS. The learned utility is then tested in another MDP, with $p=0.3$. In both environments, we evaluate agents' behaviors by estimating the probability of the desired sequence $S_0\rightarrow S_1\rightarrow g$. The result is shown in Fig.\ref{fig:prob_shift}: Although both agents with MEIP and MaxEnt-IRL behave perfectly with $p=0.1$, only the agent with MEIP still performs perfectly with $p=0.3$. The MaxEnt-IRL agent discards temporal relations in utility and only matches mean statistics $\mu(\zeta)$, thus prefers $S_0\rightarrow b_1$ in the first transition after a probability shift. 

We also conducted an empirical study to explore the boundary of generalization of both MEIP and MaxEntIRL. If $p<0.24$, the difference between MEIP and MaxEntIRL would be insignificant since the reward learned from MaxEntIRL can also discourage agents from taking $a_2$. We also notice that if $p>0.85$, which induces an extremely high probability for agents to move to $b_1$ after taking $a_1$, neither MEIP nor MaxEntIRL learns meaningful utility. 

\subsection{Experiment 2: Structural Change}
Structural change happens when the utility is transferred between two environments whose underlying PGMs for the transition $\mathcal{T}$ have the same fluent nodes $\mathcal{F}$ but different causal structures or different numbers of nodes. Note that causal structures of environments always implicitly enforce ordering in demonstration sequences. If the ordinal information in these sequences only reflects causality, there should be no difference between MEIP and IRL. However, we humans are cultural creatures. There are lots of things we do in an order not because they are the only feasible ways, but due to certain social conventions, \textit{e.g.} the traditional order in a wedding ceremony, the stroke order in hand-writing, \textit{etc}. We acquire social utility by following these conventions. It is under these situations does MEIP differentiate from IRL.

\begin{figure}[h!]
    \centering
    \includegraphics[width=0.90\linewidth]{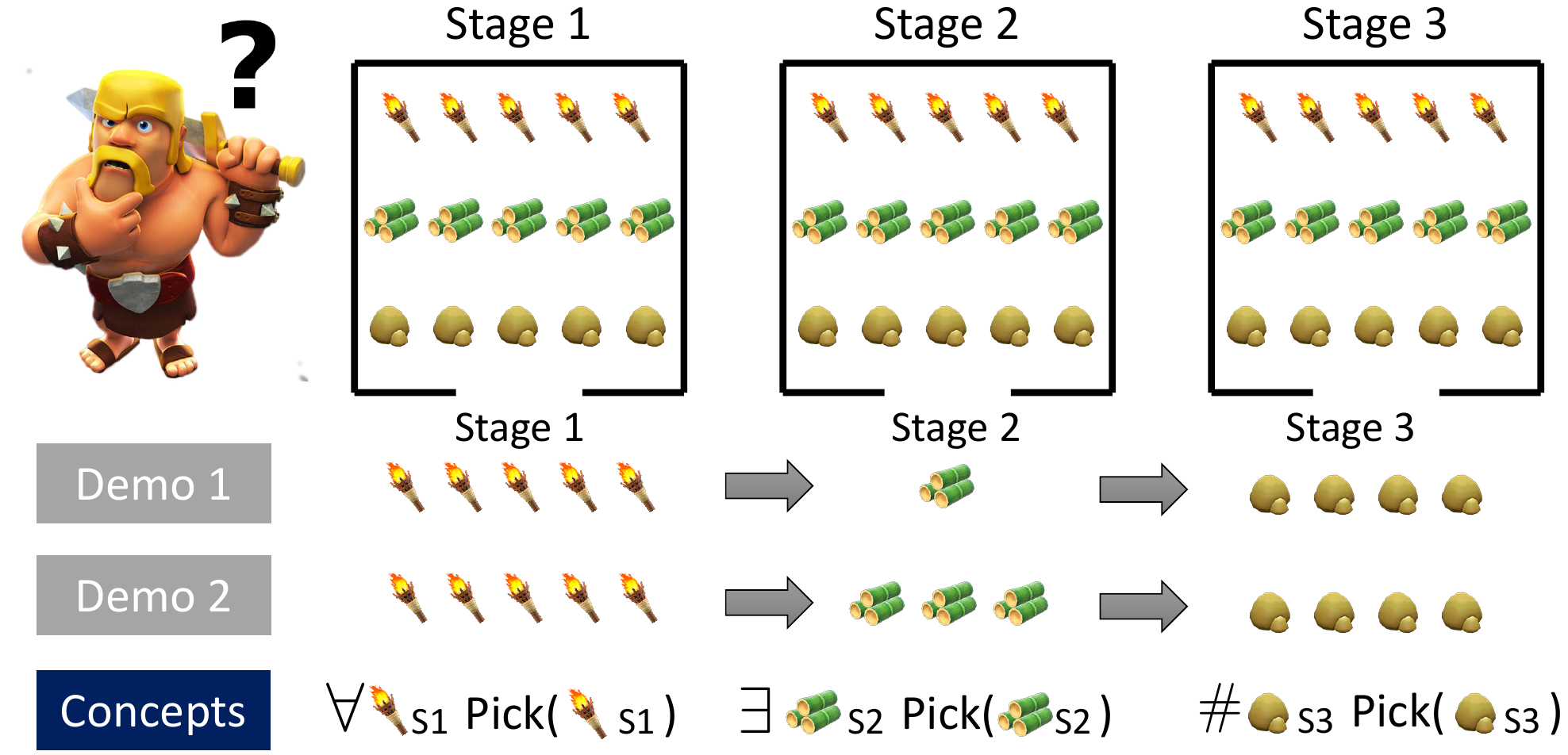}
    \caption{There are 3 stages with abundant preps. Alex was shown demos by the chief who followed a prescribed order to fetch different numbers of \textbf{torch, bamboo, clay}.}
    \label{fig:ritual}
\end{figure}

Consider the scenario in Fig.\ref{fig:ritual} that hypothetically took place in early history when utterance was not at all easy for our ancestors. Alex was a new-comer to a tribe. He was invited to a ritual host by the chief of the tribe. The ritual went like this: There were 3 stages in total. In stage 1, Alex saw the chief took out \textbf{all torches} {\includegraphics[width=0.03\linewidth]{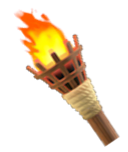}}. After stage 2, Alex was showed \textbf{1 or 3 bamboos} {\includegraphics[width=0.03\linewidth]{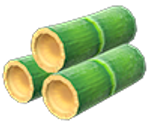}}. Eventually, the chief fetched \textbf{4 pieces of clay} {\includegraphics[width=0.03\linewidth]{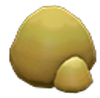}} from stage 3. As a member of this tribe, Alex need to understand the process of this ceremony. And the metric to test his understanding is how he would imagine himself hosting it, presuming some environmental changes.

\begin{figure}[t!]
    \centering    
    \begin{subfigure}[b]{0.49\linewidth}
        \includegraphics[width=\linewidth]{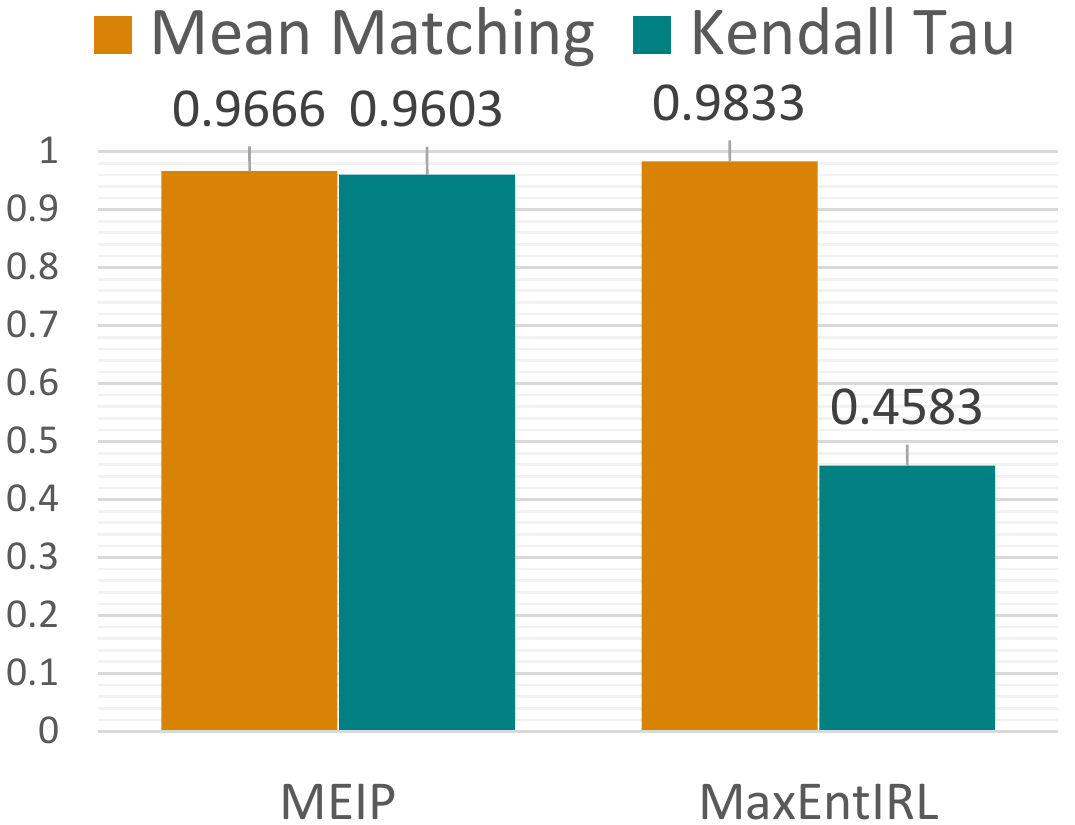}
        \caption{}
        \label{fig:result_ritual}
    \end{subfigure}
    \begin{subfigure}[b]{0.49\linewidth}
        \includegraphics[width=\linewidth]{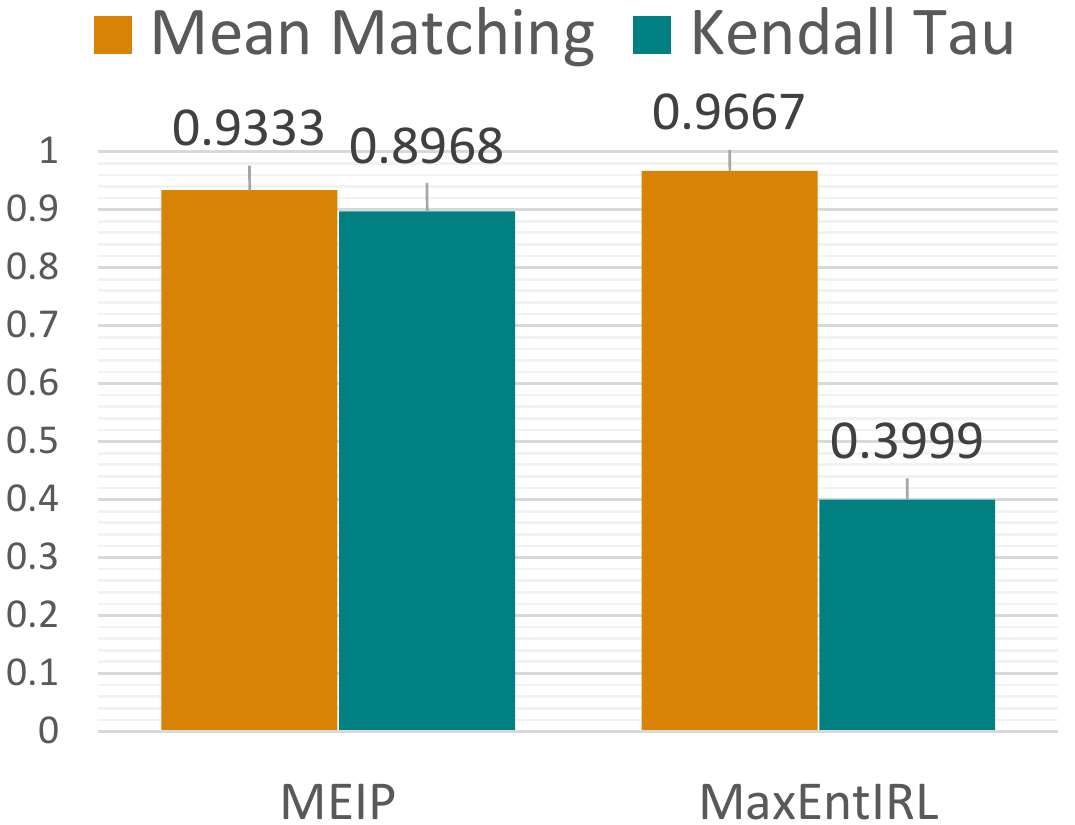}
        \caption{}
        \label{fig:result_ritual_gen}
    \end{subfigure}
    \caption{(a) Ordinal and mean matching when agents are given world models that do not enforce the demonstrated temporal ordering. (b) Results after an extra change in entity quantity. These results are the average of trajectory samples from $20$ convergences of MCTS with the learned utility.}
    \label{fig:structure}
\end{figure}

The first evaluation metric, mean matching, tests whether the learned utility is associated with the correct set of concepts and thus can be transferred to another environment with different quantities of objects. In the demonstrated setting, there are 5 torches {\includegraphics[width=0.03\linewidth]{figures/torch}}, 5 bamboos {\includegraphics[width=0.03\linewidth]{figures/bamboo}} and 5 pieces of clay {\includegraphics[width=0.03\linewidth]{figures/clay}} in each stage. After the structural change, there are 6 pieces of each objects. The ground-truth concepts of these demonstrations are $\forall$ {\includegraphics[width=0.03\linewidth]{figures/torch}}$_{S1}$ $Pick(${\includegraphics[width=0.03\linewidth]{figures/torch}}$_{S1})$ , $\exists$ {\includegraphics[width=0.03\linewidth]{figures/bamboo}}$_{S2}$ $Pick(${\includegraphics[width=0.03\linewidth]{figures/bamboo}}$_{S2})$ , $\#$ {\includegraphics[width=0.03\linewidth]{figures/clay}}$_{S3} Pick(${\includegraphics[width=0.03\linewidth]{figures/clay}}$_{S3})$. In plain English, the learner needs to fetch all torches from stage 1, any number of bamboo from stage 2 and 4 of pieces of clay from stage 3. We estimate the mean statistics of matching these concepts in sequences planned with utility learned by MEIP and MaxEnt-IRL. Both agents successfully discover the correct set of concepts with the help of our concept language (Fig.\ref{fig:result_ritual}). These concepts empower them to generalize learned utility to environments with different quantities of entities (Fig.\ref{fig:result_ritual_gen}). 

The second metric is for ordinal relations in sequences. We estimate the Kendall $\tau$ of planned sequences with the ground-truth ordering $S1\rightarrow S2\rightarrow S3$. To test this generalization, structural change needs to alter causality. And this alternation is done by controlling the world model provided to agents. When the transition in the world model directly enforces the required ordering, both agents have $\tau=1$ when planning with this world model. However, after changing to a transition without causality constraint on ordering, namely, agents can choose to enter any stage in any order, only the MEIP agent can attain the desired ordering ($\tau=96.03\%$) with the learned utility. We also tried a controlled setting where we provide a world model that does not have causality constraints to both agents. The contrast between the planned behaviors of the MEIP agents and the MaxEntIRL agent remains the same. These results justify that MEIP can learn not only correct concepts but also the desired temporal order. On the other hand, MaxEnt-IRL fails to capture the ordinal information. As illustrated in Fig.\ref{fig:structure}, mean matching does not enforce ordinal utility learning.

\begin{figure}[t!]
    \centering
    \includegraphics[width=0.80\linewidth]{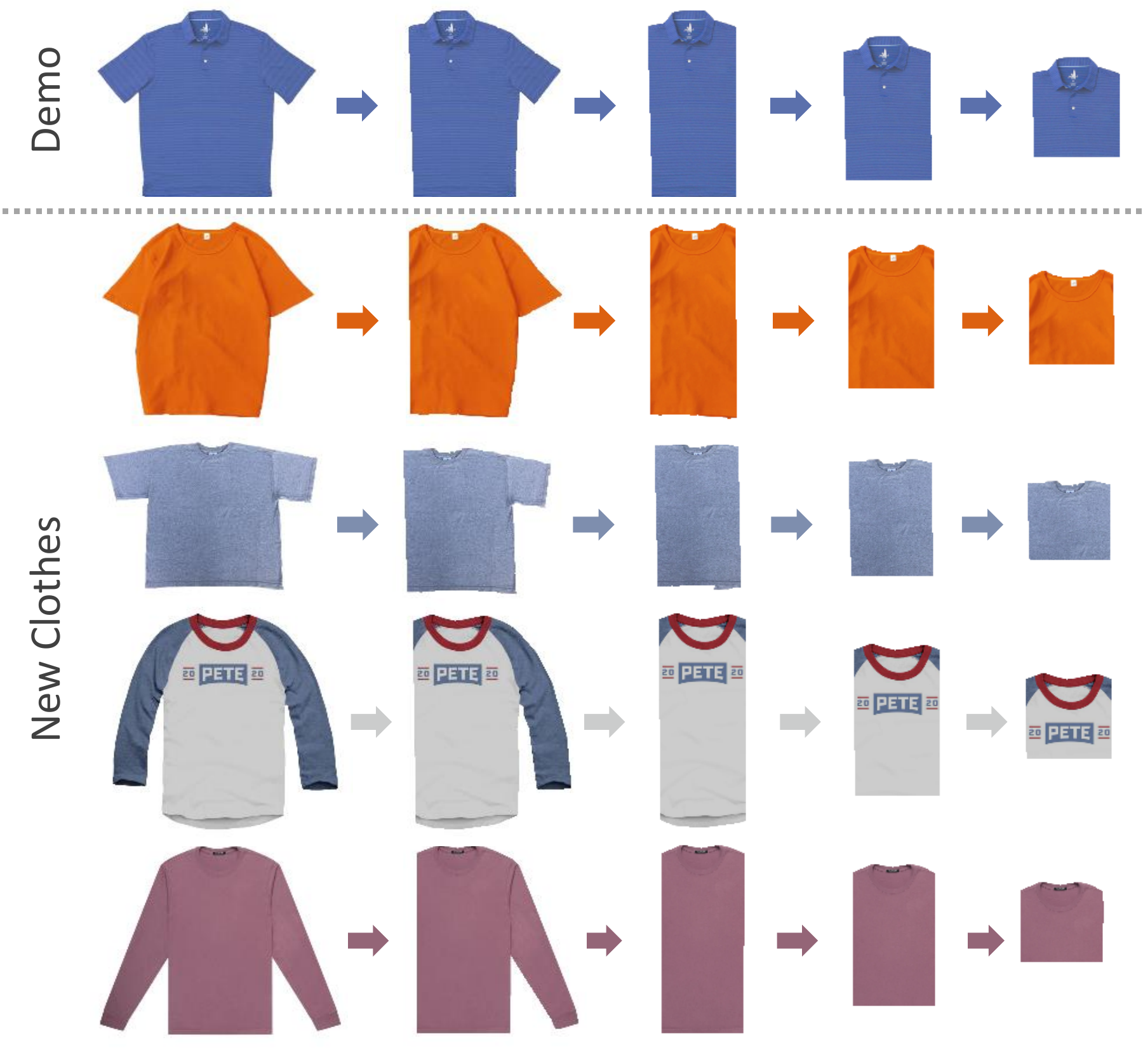}
    \caption{Qualitative results of transferring the learned utility to folding different clothes. Note that the shape of \polygon s and the number of \edge\Hquad segments in the testing clothes are substantially different from the one in demos.}
    \label{fig:result_shirts}
\end{figure}

\subsection{Experiment 3: \emph{Learning to Fold}}
Another task that reflects the ubiquity of this social utility in our daily life is the running example we start from the introduction, cloth folding. If the utility was only associated with the final state that the cloth becomes a square whose area is within a range, there would not be those exemplar sequences you recall when you hear this task. We conduct an experiment to learn the utility of cloth folding in a visually and geometrically authentic simulator.

\begin{figure}[t]
    \centering
    \includegraphics[width=0.7\linewidth]{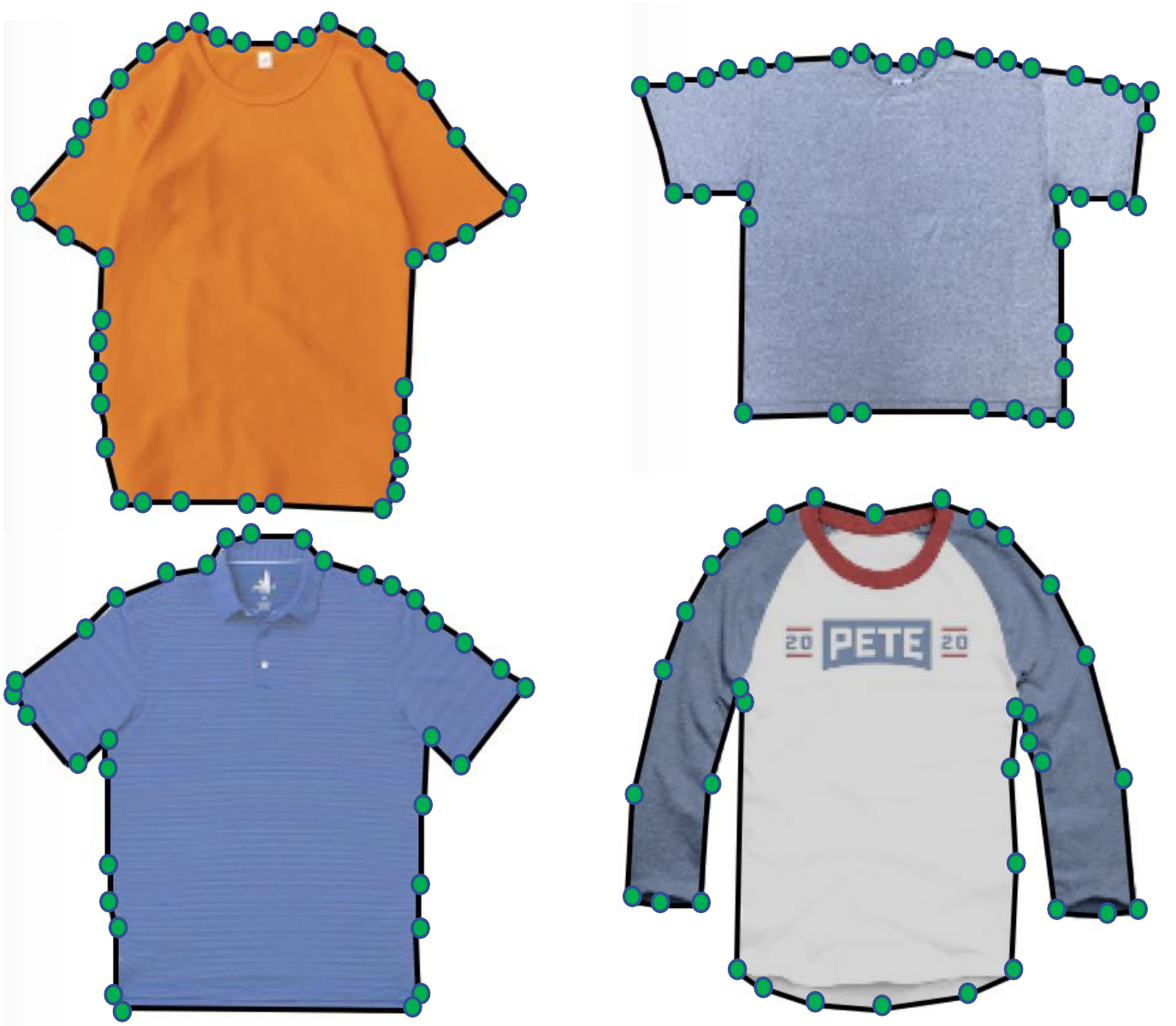}
    \caption{Nodes of the underlying structures, illustrated as \edge s and \vertex es of different clothes. They are significantly different in different clothes.}
    \label{fig:cloth_edge}
\end{figure}

To represent the cloth to be folded under our formulation, we adopt a grammar, Spatial And-Or-Graph (S-AOG) \citep{zhu2007stochastic} for the image input. Details of this grammar, as well as other technical details, can be found in the supplementary. Given the visual input of a cloth, the agent should \emph{parse} it with this grammar to acquire a first-order representation. For the sake of simplicity, we ignore the uncertainty from perception. We also endow the agent with an action model, assuming it can imagine geometric transformations like a toddler. As such, the agent has sufficient knowledge of each problem instance to rollout with MCTS. In our experiments, we collect 15 ``good folds'' from human demonstrators. When presented with these folding sequences, the agent with MEIP learns the utility and concepts that successfully generalize to unseen clothes, see Fig.\ref{fig:result_shirts}. Even though all shirts (or sweater) look similar on their appearances, their underlying structures are significantly different in the number, location, and orientation of \edge s and \vertex es, see Fig.\ref{fig:cloth_edge}. 

\section{Concluding Remarks}
In this work we propose a new quest for learning generalizable task representation, especially its utility, from demonstrations. This problem lies outside of the regime of inverse reinforcement learning and thus dubbed \emph{Generalized Inverse Planning}. We then outline \emph{the computational principles in the cognitive process} \cite{marr1976understanding} of lifted non-Markovian utility learning, which we model as Maximum Entropy Inverse Planning (MEIP). Comparing to existing inverse reinforcement learning methods, our model learns a task representation that generalizes regardless of probability shift and structural change in the environment. To highlight this contribution, we exclude irrelevant representation learning by adopting classical assumptions and representation in planning, \textit{i.e.} grounded semantics of entities and relations are given \textit{a priori}, as well as an action model over them. This kind of assumption was also made in original works of IRL \cite{abbeel2004apprenticeship}, \cite{ratliff2006maximum},\cite{ziebart2008maximum}, \cite{munzer2015inverse}. To disclaim, we are aware that such assumption on priors might be regarded as too strong in modern days since there are some recent progress in deep reinforcement learning that successfully induces them with a weaker assumption of relational inductive biases \cite{zambaldi2018deep}. 
Our model is general enough to be extended to those neural networks. It would an interesting future work to investigate their synergy. 

\section{ Acknowledgments}
The authors thank Prof. Ying Nian Wu and Dr. Yixin Zhu of UCLA Statistics Department, Prof. Guy Van den Broeck of UCLA Computer Science Department for useful discussions. The work reported herein was supported by ONR MURI grant N00014-16-1-2007, ONR N00014-19-1-2153, and DARPA XAI N66001-17-2-4029.
\bibliography{reference}
\newpage
\newpage

\section{Algorithmic Details}
\subsection{\ac{meip}}
For the purpose of generalization, the ranking function $g_\omega(x)$ of utility is designed to be a piece-wise linear function. When $x\in \mathbb{R}^1$, the form of the function is:
\begin{equation*}
    g(x) = \begin{cases}
    k_0 x + b_0 & x \in [x_0, x_1] \\
    k_1 x + b_1 & x \in [x_1, x_2] \\
    ... \\
    k_n x + b_n & x \in [x_n, x_{n+1}]. 
    \end{cases}
    \label{eq:piecewise_linear}
\end{equation*}
Cases where $x$ has higher dimensions can be derived accordingly. To make sure the continuity, we further add constraint on the connection between two consecutive bins, i.e. $k_i x + b_i = k_{i+1} x + b_{i+1}$. 

The training process of \ac{meip} follows the principle of \emph{analysis by synthesis}. According to Eq.11 there are two parts to be optimized, one for demonstrations $\Psi^E$ and one for sampled plans $\Psi$. Specifically, the learning is implemented with a Ranking SVM, see Eq.12. Ordinal relations between pairs are specified in the main text after Eq.11. For a $\prec$ pair $\langle s_1, s_2 \rangle$, we set $g(s_2) - g(s_1) \geq 1 - \xi$ for the Ranking SVM. 

The sampling process is based on a Monte Carlo Tree Search (MCTS). First, we initialize the value function $V$ of the MCTS. We keep doing rollout to compute $\tau_\omega(\zeta)$ for trajectories $\zeta$ until value functions in MCTS converges. After that, we choose to sample a small number of trajectories $\Psi$ according to the values on the branches of the Monte Carlo Tree. These trajectories, together with the human demonstrations $\Psi^E$, are used to update the utility function. 

\subsection{Concept Pursuit}
As mentioned in the main text, we generate first-order concepts from a concept language. However, the size of this language could be infinitely large. Thus we adopt the principle of Occam's razors. We herein describe the algorithm for concept pursuit in \autoref{alg:pursuit}. 

There are three main steps in the pursuit process. First, we sort the concepts according to the complexity. The measurement of complexity is based on the number of predicates (or functions) in a concept. The more predicates (functions) in a concept, the higher complexity it is. This discrete nature introduces \emph{levels} in the complexity.

\begin{algorithm}[h!]
\textbf{Input:} Sorted Concept Set $\mathcal{C}$; Threshold $\epsilon$ \\
\KwData{The Set of State Pairs : $\{\langle s_i, s_j \rangle, ...\}$}
\KwResult{Selected Concept Set $\Delta$; Optimized Parameters $\omega$}
\textbf{Init:} Selected Concepts Set $\Delta = \{\}$\\
\ForEach {concept subset $\mathcal{C}_{l_i} \subset \mathcal{C}$}{
    \ForEach {first-order concept $c \in \mathcal{C}_{l_i}$}
    {
        violation count $v_c \gets 0$ \\
        \ForEach {state pair $\langle s_i, s_j \rangle$}
        {
            \If{$c(s_j) \neq c(s_i)$ }{$v_c \gets v_c+1$}
        }
        \If{$v_c = 0$}
        {
            remove $c$ from $\mathcal{C}_{l_i}$
        }
    }
}

\ForEach{concept subset $\mathcal{C}_{l_i} \subset \mathcal{C}$}{
    \While {True} {
        $\{\delta_{tmp}\} = \{\}$ \\
        \ForEach{$c \in \mathcal{C}_{l_i}$}{
            compute $\delta_{c}$ by calling \autoref{alg:meip} with $\Delta + c$\\
            add $\delta_{c}$ to $\{\delta_{tmp}\}$
        }
        \If {$max(\{\delta_{tmp}\}) \leq \epsilon$}{
            break
        }
        \Else {
            $c_{max} = argmax(\{\delta_{tmp}\})$ \\
            $\Delta \gets \Delta + c_{max}$ \\
            update $\omega$
        }
    }
}
\caption{Concept Pursuit}
\label{alg:pursuit}
\end{algorithm}

The second step is to exclude concepts that are not relevant at all. We can select those relevant concepts by simply counting the number of violations in the state pairs. The last and most important step is to select concepts according to Eq. 19 for the utility we learn. Each concept is added to the selected set $\Delta$ in a step-wise greedy manner. Note that concepts that share the same predicate (or function) and the same domain but different quantifiers are mutually exclusive.

\section{Details for Experiment 2}
\subsection{Concept Space in Ritual Learning}
In the ritual learning experiment, we only consider the concepts that are generated by the rule $QPD$ in which $Q$ is the quantifier, $P$ is the predicate, and $D$ is the domain. In the ritual learning experiment, the domain $D=(O, S)$ in which $O$ is the set of object types and $S$ is the set of stages. A primitive concept in this experiment can be expressed as:
\begin{equation*}
    \forall|\exists|\# \texttt{picked} (o\in O, s \in S)
    \label{eq:ritual_concept}
\end{equation*}

The \emph{primitive predicate}, \texttt{picked}, describes whether the host carries some type of object from a specific stage. As introduced in the main text, \emph{primitive concepts} represented by \emph{atomic formulas} with the primitive predicate are \emph{terms} for formulas of more complex concepts. Here these more complex concepts are encodings of the interrelation between primitive concepts.

\subsection{Experimental Details}
\paragraph{The Environment} is designed to be an analogy to a ritual. There are different stages in the environment. Both the agent and the demonstrator are required to choose a certain stage before they can advance to pick up objects in it. After locking down a specific stage, one will be asked to choose one type of object and pick up the $[0,+\infty)$ of them ($+\infty$ means all). None of the stages can be visited more than once. The ritual will be terminated after all stages are visited.

\paragraph{Human Demonstrations} consist of at least 3-5 sequences. A sequence consists of an ordered descriptions of objects that the demonstrator obtains at each stage. The demonstrator can only choose to pick one type of object at one stage without limitation on the quantity. Examples of demonstrations are $5\times$ {\includegraphics[width=0.03\linewidth]{figures/torch}}$_{S1}$ $\rightarrow$ $3\times$ {\includegraphics[width=0.03\linewidth]{figures/bamboo}}$_{S2}$ $\rightarrow$ $4\times$ {\includegraphics[width=0.03\linewidth]{figures/clay}}$_{S3}$ or $1\times$ {\includegraphics[width=0.03\linewidth]{figures/bamboo}}$_{S1}$ $\rightarrow$ $2\times$ {\includegraphics[width=0.03\linewidth]{figures/clay}}$_{S2}$ $\rightarrow$ $2\times$ {\includegraphics[width=0.03\linewidth]{figures/torch}}$_{S3}$. There is no doubt that we can have a longer demo if it is legal in the environment, although each demo only have 3 stages in our experiment. Note that all demos must contain exactly the same set of specific concepts. 

\paragraph{Hyper-parameters} of this experiment are listed as the following. MCTS converge condition: terminate after 3000 iterations. Size of sample trajectories $\Psi=\{\zeta_j\}$: 5. Upper confident bound coefficient: 1. 

\section{Details for Experiment 3}
\subsection{Representation for Visual Input of Clothes}
We adopt a stochastic grammar, Spatial And-Or-Graph (S-AOG) \cite{zhu2007stochastic} for the visual input of all clothes. The design of this grammar follows Gestalt Laws in vision \cite{marr1982vision}. Here we informally summarize its \emph{production rules}. 
A \texttt{cloth} is an \emph{And} node that \emph{produces} a set of \polygon s. 
The number of \polygon s may change after being folded. Since some \polygon s may be occluded in the visual input, we adopt a 2.1D representation \cite{nitzberg19902}. The 2.1D representation is a layer representation. In our case, the order of the layer is consistent with the folding order. All \polygon s belong to the same class with a template set of \emph{fluents}. They produce a set of line segments, \edge s. Different configurations of \edge s in one {\polygon} consist an \emph{Or} node. All \edge s also belong to the same class. Each {\edge} is associated with two \vertex es as its attribute. Each \vertex is specified by its coordinate. Classes introduced above are regarded as domains in the concept language. 

The full fluent set for this grammar is designed following axioms in Euclidean geometry, as showed in \autoref{tab:folding_function}. Edge s, vertices, surfaces and their relations are also believed to be our core knowledge developed in early age \cite{spelke2007core}. Fluents for \edge s are categorized into functions/relations between \edge s \textit{e.g.} parallel, and functions/relations between one \edge and one {\vertex} of another \edge \textit{e.g.} distance. Other fluents, such as \texttt{Logo} and \texttt{Neck}, are for \polygon s, which are visual features. With these classes and fluents, we can generate \emph{concepts} from the concept language. 

\begin{figure}[h]
    \centering
    \includegraphics[width=\linewidth]{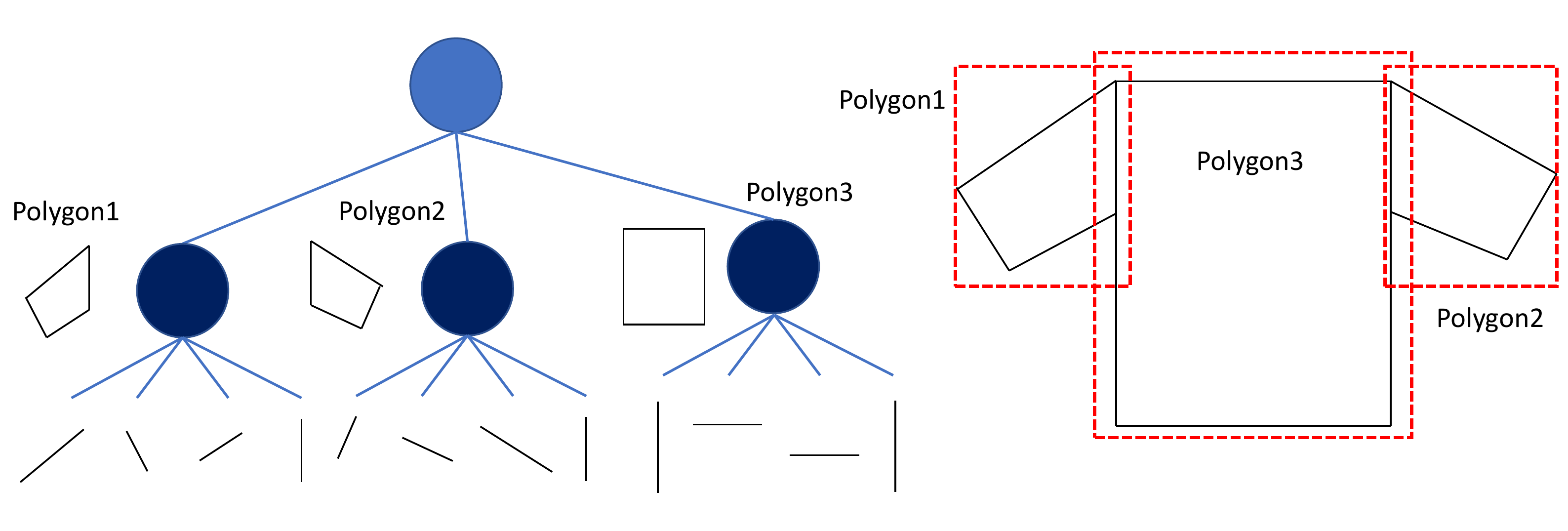}
    \caption{A minimal parse graph of the S-AOG}
    \label{fig:cloth_pg}
\end{figure}

During planning, the visual input of the cloth from each situation is parsed into a \emph{parse graph} ($pg$) of the S-AOG. As shown in \autoref{fig:cloth_pg}, $pg$ is a hierarchical representation in which the terminal node is an \edge with two \vertex es and non-terminal nodes are \polygon s. Clothes with the simplest structures, such as a shirt, are initially parsed into three polygons and their own affiliated edges. 

\begin{table}[t]
    \centering
    \begin{tabular}{|c|c|c|} \hline
        Name & Arity & Type \\ \hline
        \texttt{Edge Length} & Unary & function \\ \hline
        \texttt{Logo} & Unary & function \\ \hline
        \texttt{Neck} & Unary & function \\ \hline
        \texttt{Vt2Vt Distance} & Binary & function \\ \hline
        \texttt{Vt2Edge Distance} & Binary & function \\ \hline
        \texttt{Parallel} & Binary & predicate \\ \hline
        \texttt{Perpendicular} & Binary & predicate \\ \hline
        \texttt{Vertex on Edge} & Binary & predicate \\ \hline
        \texttt{Edge on Edge} & Binary & predicate \\ \hline
        \texttt{Vertex in Polygon} & Binary & predicate \\ \hline
    \end{tabular}
    \caption{Fluent space for learning to fold}
    \label{tab:folding_function}
\end{table}

\subsection{Experimental Details}
\begin{figure}[h!]
    \centering
    \includegraphics[width=\linewidth]{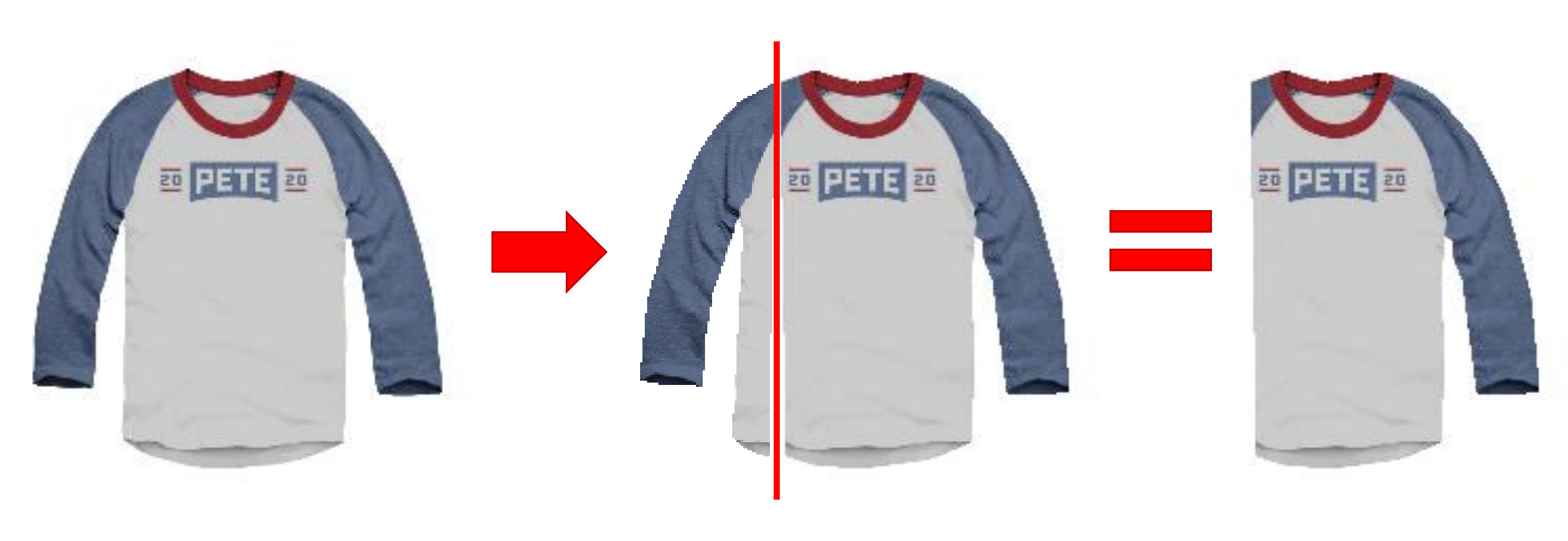}
    \caption{The action is a folding line in the simulator}
    \label{fig:folding_line}
\end{figure}
\paragraph{The Environment} is a simulator for folding. The demonstrator is asked to draw a folding line that splits a polygon into two new polygons. In \autoref{fig:folding_line}, we show an example of a legal folding line in the simulator. The polygon is separated by the folding line and the small one will be flipped to the back of the larger one after each fold.

\paragraph{Human Demonstrations} are given by folding a demo shirt (or sweater). In a demo, states of the shirt are recorded, serialized as a sequence. A fold is not reversible therefore the demonstrator needs to consider the final state at every step. If the demonstrator made a ``bad fold", it could have a significant impact on the final state. Unlike the previous experiment, the demonstrator does not have a prescribed concept set during the demo process. Instead, the demonstrator needs to conduct a folding sequence that can lead to good final states which meet their own criteria of ``good folds" based on \emph{default} sequences in their real-life habits. We collected 15 sequences as demonstrations. 

\begin{figure}[h!]
    \centering    
    \begin{subfigure}[b]{0.52\linewidth}
        \includegraphics[width=\linewidth]{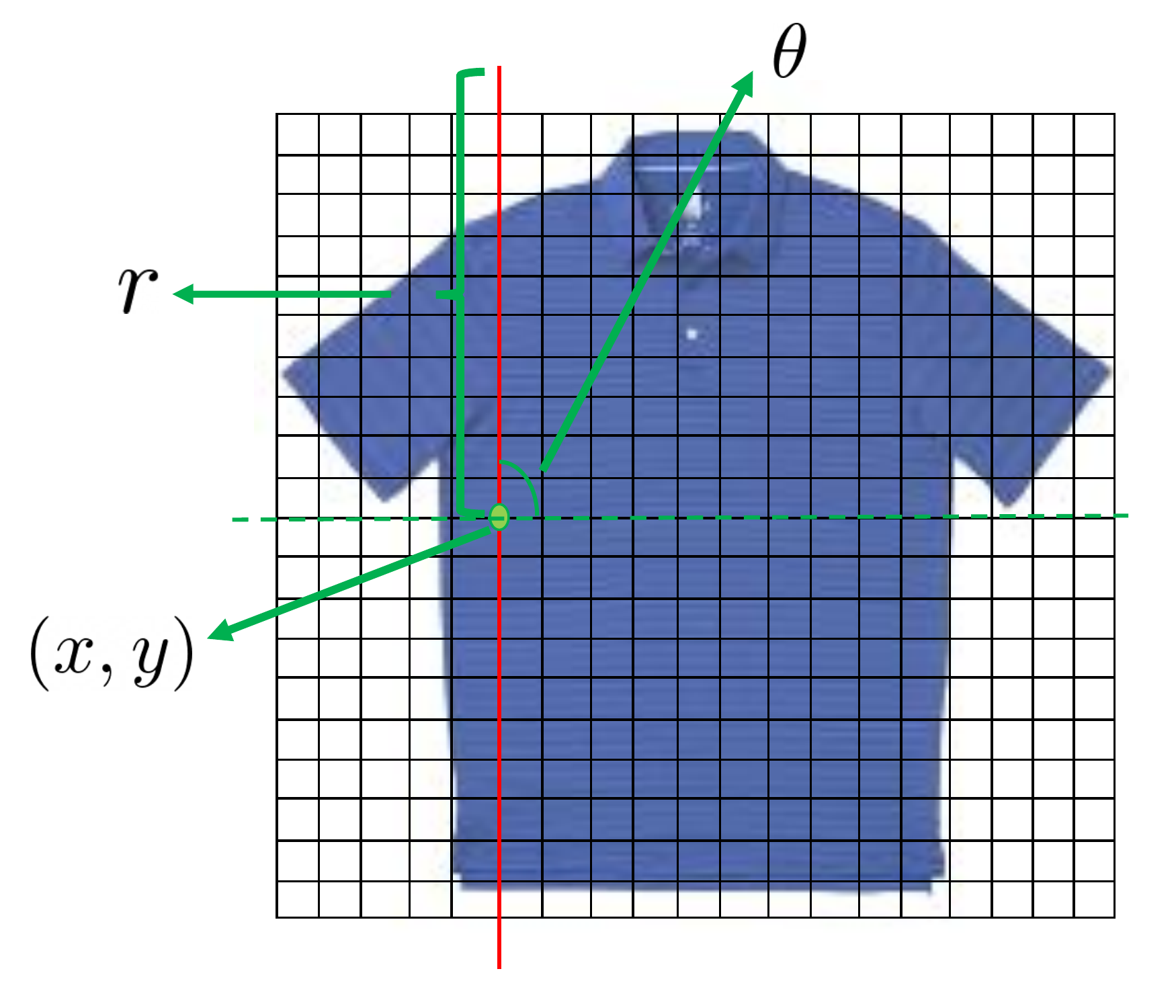}
        \caption{}
        \label{fig:cloth_action}
    \end{subfigure}
    \begin{subfigure}[b]{0.46\linewidth}
        \includegraphics[width=\linewidth]{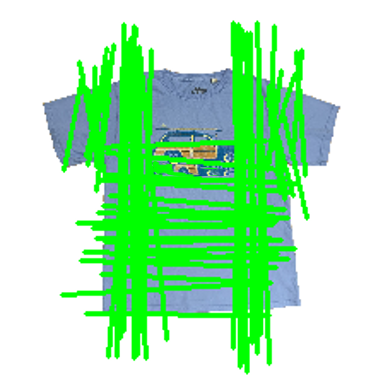}
        \caption{}
        \label{fig:action_proposal}
    \end{subfigure}
    \caption{(a) The discretized action space for folds. (b) Samples of folds that with high probabilities.}
    \label{fig:action_space}
\end{figure}

\paragraph{The Action Space} of a folding line is discretized to make MCTS applicable. It is defined as a tuple $a=(x,y,r,\theta)$ and each parameter is discretized accordingly. As illustrated in \autoref{fig:cloth_action}, $(x, y)$ is coordinate of a point on the grid. The grid is a discretization of the bounding box of a shirt. $r$ is the radius of the folding line and $\theta$ is the angle. Note that some folding lines may be redundant, therefore we need to check the uniqueness of each folding line. 

Even though we have discretized action space, it is yet too large for MCTS with limited computational resources. Thus, it is necessary to have a reasonable number of legal folds for the MCTS. The solution is to learn a probabilistic distribution over the action space. The folds that are similar to demo folds will be associated with high probabilities. We assume that each parameter in $a$ follows a normal distribution around some exemplars. See \autoref{fig:action_proposal}. 

\end{document}